\pdfoutput=1

\documentclass{article}
\usepackage{ijcai25}

\usepackage{times}
\usepackage{soul}
\usepackage{url}
\usepackage[hidelinks]{hyperref}
\usepackage[utf8]{inputenc}
\usepackage[small]{caption}
\usepackage{graphicx}
\usepackage{amsmath}
\usepackage{amsthm}
\usepackage{booktabs}
\usepackage{algorithm}
\usepackage{algorithmic}
\usepackage[switch]{lineno}

\usepackage[nameinlink,capitalize]{cleveref}

\usepackage{comment}
\usepackage{xcolor}
\definecolor{blue}{rgb}{0, 0, 1}

\usepackage{amsmath}
\usepackage{amssymb}
\usepackage{mathtools}
\DeclareMathOperator*{\argmax}{arg\,max}

\usepackage{pifont}

\usepackage{multirow}
\usepackage{array}
\usepackage{makecell}

\usepackage{subcaption}

\usepackage{xcolor}

\usepackage{booktabs}

\definecolor{orange}{HTML}{FF7F0E}
\definecolor{dodgerblue}{HTML}{1F77B4}
\definecolor{mplgreen}{HTML}{2CA02C}

\title{Flexible Blood Glucose Control:\\ Offline Reinforcement Learning from Human Feedback}

\author{
Harry Emerson$^1$
\and
Sam Gordon James$^1$\and
Matthew Guy$^{1,2}$\And
Ryan McConville$^1$\\
\affiliations
$^1$University of Bristol, United Kingdom\\
$^2$University Hospital Southampton, United Kingdom\\
\emails
\{harry.emerson, sam.james, ryan.mcconville\}@bristol.ac.uk,
matthew.guy@uhs.nhs.uk
}

\begin{document}

\maketitle

\begin{abstract}
     Reinforcement learning (RL) has demonstrated success in automating insulin dosing in simulated type 1 diabetes (T1D) patients but is currently unable to incorporate patient expertise and preference. This work introduces PAINT (Preference Adaptation for INsulin control in T1D), an original RL framework for learning flexible insulin dosing policies from patient records. PAINT employs a sketch-based approach for reward learning, where past data is annotated with a continuous reward signal to reflect patient's desired outcomes. Labelled data trains a reward model, informing the actions of a novel safety-constrained offline RL algorithm, designed to restrict actions to a safe strategy and enable preference tuning via a sliding scale. In-silico evaluation shows PAINT achieves common glucose goals through simple labelling of desired states, reducing glycaemic risk by 15\% over a commercial benchmark. Action labelling can also be used to incorporate patient expertise, demonstrating an ability to pre-empt meals (+10\% time-in-range post-meal) and address certain device errors (-1.6\% variance post-error) with patient guidance. These results hold under realistic conditions, including limited samples, labelling errors, and intra-patient variability. This work illustrates PAINT's potential in real-world T1D management and more broadly any tasks requiring rapid and precise preference learning under safety constraints.
\end{abstract}

\section{Introduction}

Glucose controllers automatically regulate insulin dosing to respond to changes in blood glucose levels. Reinforcement learning (RL) has shown promise in blood glucose management, achieving state-of-the-art results in virtual type 1 diabetes (T1D) patients \cite{Tejedor2020ReinforcementReviewc,Emerson2023OfflineDiabetesc,Hettiarachchi2024G2P2CDiabetes,Jaloli2023Long-TermNetwork}. Despite their strong performance in simulation, RL glucose controllers remain unsuitable for real-world use, lacking certain functionalities of simpler rule-based systems.

One such limitation is the inability of RL controllers to integrate patient feedback and expertise \cite{Tejedor2020ReinforcementReviewc,Zhu2021DeepReview}. T1D management is highly individualised, with optimal strategies shaped by each person’s unique lifestyle and physiology \cite{Redondo2023HeterogeneityMellitus}. Patients and their carers spend years learning to manage their condition, and leveraging this knowledge could enhance glucose control. Current commercial controllers allow parameter adjustments to support personalised goals, such as event-specific control or secondary metric optimisation \cite{Hartnell2021Closed-loopGuideb,Berg2024GoalDiabetes}. However, effective customisation often demands expertise and trial-and-error testing, which may exceed patients’ capabilities. Furthermore, most controllers are constrained by sensor limitations, despite efforts to integrate wearables that monitor broader glucose influences \cite{Daskalaki2022TheSurvey}. Patients actively managing T1D are inherently aware of these external factors and could provide valuable context to improve decision-making.

This work presents, \textbf{PAINT} (\textbf{P}reference \textbf{A}daptation for \textbf{IN}sulin Control in \textbf{T}1D), a novel approach for training safe and flexible RL policies from pre-collected patient data. PAINT is comprised of two core components: a sketch-based tool for patient preference elicitation, and a safety-constrained offline RL controller. Patients convey preference information by highlighting beneficial dosing strategies in their historical data. This is performed by drawing a continuous reward signal, indicating the proximity of their current strategy to a goal state. A reward model is then trained using the preference data to approximate the patient's reward function. The reward labelled data tunes the novel offline RL algorithm, which constrains controller actions to a verifiably safe strategy, ensuring preferences do not elicit dangerous behaviours. The constraint can be modified by the patient using a sliding scale, allowing greater precision in modifying the strength of the preference. 

Extensive evaluation shows PAINT is able to replicate commercial controller functionality while adopting safer and more effective dosing strategies, reducing patient risk by 15\% across common blood glucose goals. Unlike current commercial blood glucose controllers, PAINT doesn't require users to know how to achieve their goals. Patients can simply specify their desired outcome via the sketching the tool, and PAINT determines the optimal way to achieve it. If patients do wish to include their expertise, PAINT also enables feedback on individual actions. The incorporation of patient expertise was explored across two case studies, with the goal of improving insulin dosing before meals and better responding to device errors. With the inclusion of patient expertise, PAINT demonstrated a 10\% increase in healthy post-meal blood glucose levels and a 1.6\% reduction in variance after device errors. PAINT also shows strong robustness to real-world challenges, achieving competitive results with as little as two days of reward-labelled samples, while effectively handling labelling errors and intra-patient diversity. 

This work represents the first RL approach for incorporating patient expertise and preferences in T1D management and highlights its potential as a component in real-world glucose controllers. PAINT is likely to be applicable to any domain requiring precise user-adaptive control of RL polices subject to pre-defined constraint. 

\section{Related Work}

\subsubsection{Reinforcement Learning in Diabetes Management}

Prior research in RL and T1D has predominantly focussed on incremental performance improvements through architectural changes, validated exclusively in simulation \cite{Jaloli2024Basal-bolusMethodology,Hettiarachchi2024G2P2CDiabetes,Yu2023ARLPE:Diabetics}. In contrast, less research has focused on addressing practical challenges in RL's application to real-world management. The application of offline RL enabled risk-free training in T1D; using pre-collected datasets created under verifiably safe policies to train RL controllers \cite{Emerson2023OfflineDiabetesc,Beolet2023End-to-endControl}. Off-policy evaluation has created a promising avenue for the safe evaluation of novel strategies \cite{Beolet2023End-to-endControl,Viroonluecha2023EvaluationDiabetes}. Works in interpretability have contributed to improved trust in RL decision-making and enabling patient intervention in RL actions \cite{Melloni2024InterpretingExplanations,Lim2021AValidationb}. Controller flexibility represents an under-explored area, despite being a common feature in almost all non-RL based algorithms \cite{Wilmot2019OpenStudy}. To the best of the authors' knowledge, no existing RL-based glucose controllers allow policy adjustment for patient preference or to integrate their expertise.

\subsubsection{Learning from Human Feedback}

Human feedback is critical for aligning RL agents with real-world objectives across diverse domains \cite{Casper2023OpenFeedback,Kaufmann2023AFeedback}. Preference-based learning has been utilised to address challenges with traditional human feedback, learning policies from reward functions inferred from preference data \cite{Metcalf2024Sample-EfficientRewardsb}. Pairwise comparisons represents one of the most common preference elicitation methods, requiring human labellers to choose their preferred example from a pair of state-action trajectories, based on alignment with their desired goals \cite{Christiano2017DeepPreferences}. While simple, this method lacks the expressiveness for precise feedback, making it incompatible over long-horizon tasks, such as in T1D management \cite{Casper2023OpenFeedback}. In contrast, scalar labelling allows for more expressive feedback by assigning numeric values to samples to indicate preference strength \cite{Wilde2022LearningFeedback}. Most similar to this work \citeauthor{Cabi2019ScalingLearning} introduced reward sketching, where labellers draw a continuous line reflecting a robot's proximity to a goal state \cite{Cabi2019ScalingLearning}. This work adapts reward sketching for T1D, substituting video data with pre-collected diabetes data. PAINT also utilises a simplified reward cloning loss and stratified reward sampling, experimentally observing this to enable greater precision in reward learning and facilitates better performance for discrete reward signals.  

\subsubsection{Safe Reinforcement Learning}

Policies must remain safe even under patient preferences, as incorrect actions hold high risk in T1D. Safe RL is concerned with providing safety guarantees for the real-world deployment of RL agents \cite{Gu2024AApplications}. Most approaches employ constrained optimisation techniques \cite{Garcia2015ALearning}, such as Lagrangian methods \cite{Stooke2020ResponsiveMethods,As2022ConstrainedModels}. In contrast, the safe offline RL setting has been comparatively less well explored \cite{Lin2023SafeConstraints}, especially in the context of safety and human feedback \cite{Gong2024OfflineClassification}. This work presents a novel safe offline RL approach for balancing preference constraints with policy safety, modifying the offline RL algorithm TD3+BC to enable precise fine-tuning of preference strength.

\section{Methods}

\subsection{Markov Decision Process}

In the standard RL formulation, environments are characterised by a Markov Decision Process (MDP), defined by $\langle S, A, r, T, \gamma \rangle$, where $s \in S$ and $a \in A$ denote the state and action, $r(s, a)$ denotes the reward function, $T(s'|s,a)$ denotes the transition probability and $\gamma$ denotes the discount factor \cite{Sutton1998ReinforcementIntroduction}. The RL objective is then to find an optimal policy $\pi(a|s)$, which maximises $\mathbb{E}_{s_0 \sim \mu_0} V^{\pi} (s_0)$, where
\(
    V^{\pi}(s) = \mathbb{E} \left[ \sum^{\infty}_{t=0}  \gamma^t r(s_t, a_t) | s = s_0 \right],
\)
is the value function defining the expected cumulative reward under a given policy in the MDP and $\mu_0$ is the initial state distribution. 

In the offline RL setting, training access to the MDP is not assumed and instead RL agents learn from a static dataset, $D = \{ \tau_0, ..., \tau_N \}$, composed of $N$ trajectories containing $M$ contiguous tuples. $\tau_i = \{(s_{i,0}, a_{i,0}, s_{i,0}'),..., (s_{i,M}, a_{i,M}, s_{i,M}')\}$ collected under a set of unknown policies \cite{Shin2023BenchmarksLearningb}. This approach is essential for safety-critical tasks, such as blood glucose management, as the deployment of partially trained policies can be harmful \cite{Levine2020OfflineProblemsb}.

Similarly, preference-based reward learning (PBRL) is necessary to adapt policies to the dynamic and evolving needs of people with T1D. In this context, RL agents do not have access to the ground truth reward function, $r(s,a)$, but can query an expert for \textit{preference feedback} on pre-collected trajectory data \cite{Gao2024HindsightLearning}. These preferences are then used to approximate the reward function $\hat{r}(s, a)$, as to generate reward labels for the wider dataset and inform the RL agent's actions.

\subsection{Reward Sketching for Preference Labelling}

\begin{figure}[t!]
    \centering
    \includegraphics[width=\linewidth]{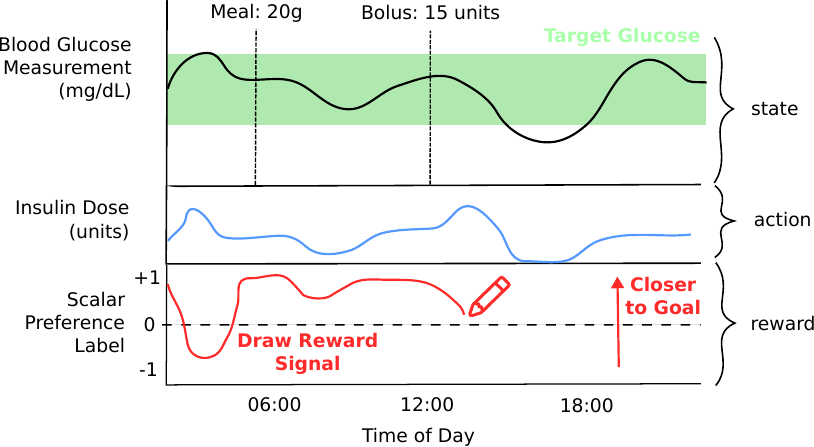}
    \caption{Overview of reward sketching for T1D. A blood glucose profile (top) with meal insulin doses and consumed carbohydrates. Recorded basal inulin doses (middle) controlled by the insulin dosing device. Participant supplied reward labelling (bottom), specifying how close the dosing behaviour is to the desired goal at a given time. Participant draws a continuous reward signal under the historical data, highlighting desirable actions and states.}
    \label{fig:reward_sketching}
\end{figure}

Despite the widespread use of pairwise comparison methods for preference elicitation, the temporally extended nature of insulin dosing actions in T1D management obscures the exact state-action pairs being rewarded, leading to suboptimal agent performance. In contrast, scalar reward labelling methods, in which labellers assign a numeric reward value to individual trajectories, have been observed to be significantly more expressive \cite{Wilde2022LearningFeedback}. For this reason, PAINT adapts the scalar reward labelling technique, \textit{reward sketching} for T1D \cite{Cabi2019ScalingLearning}. Creating an accurate and sample-efficient strategy for modify RL-based blood glucose controller policies, as illustrated in \autoref{fig:reward_sketching}. 

This modified approach presents pre-collected trajectories to patients in the form of a blood glucose profile; the standard representation for reviewing insulin dosing in T1D management. Each training sample represents a single timestep in a blood glucose trajectory, composed of a reported blood glucose, insulin, and carbohydrate value. To incorporate preferences, patients label the relevant samples with a continuous reward signal $r_t$, expressing how close the controller is to their idealised states or actions. A neural network $\hat{r}_\psi(s_t, a_t)$, parametrised by $\psi$, is then trained to minimise the mean squared error loss between the predicted and true reward labels
\begin{equation}
L(\hat{r}_\psi) = \mathop{\mathbb{E}}_{{(s_i, a_i, r_i) \sim D}} \left[ \left( \hat{r}_\psi(s_i, a_i) - r_i \right)^2 \right].  
\end{equation}
The trained model then labels the wider training dataset. Training mini-batches were generated using a stratified reward sampling procedure, ensuring that each batch contained an equal samples from one of $k$ uniformly-spaced, non-overlapping reward strata. This was performed as it was observed empirically to improve performance when trained on discrete or imbalanced reward data. 

\begin{figure*}[thbp!]
    \hspace{-5mm}
    \centering
    \includegraphics[width=0.95\linewidth]{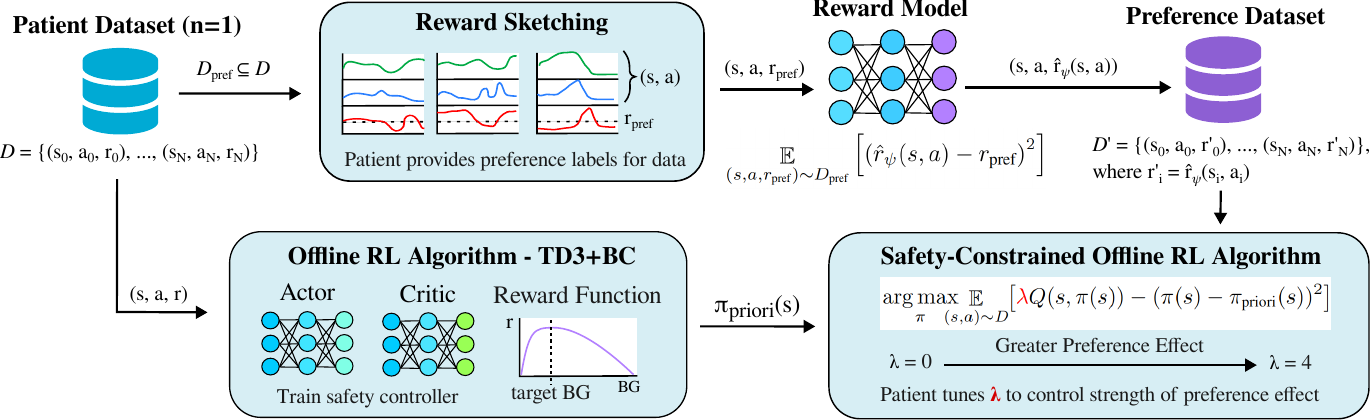}
    \caption{Full training pipeline for PAINT controller, showing the preference labelling procedure (top), and the offline RL training procedure (bottom). Users label a subset of their historical data, $D_{\text{pref}}$ in order to adapt the insulin dosing strategy to their individual needs. Reward labels, $r_{\text{pref}}$ are used to train a reward model, $r_\psi$, which is then used to label the full patient training dataset, $D'$. A generic policy, $\pi_{\text{priori}}$ is trained using a verifiably-safe reward function. $\pi_{\text{priori}}$ is then tuned using $D'$ to incorporate the user's preferences. The strength of the preference effect can be controlled via $\lambda$.}
\label{fig:full_algorithm_overview}
\end{figure*}

\subsection{Safety-Constrained Offline Reinforcement Learning}
\label{sec:safety_orl}

To ensure patient safety, PAINT utilises a \textit{safety-constrained} offline RL controller, modifying the popular TD3+BC approach, \cite{Fujimoto2021ALearningc}, allowing patients to fine-tune the strength of their preference with respect to a verifiably safe control strategy. 

TD3+BC was chosen as the basis for the controller, due to its high sample efficiency and relatively few hyperparameters \cite{Beeson2022ImprovingFine-Tuning}, enabling accurate dosing decisions with fewer samples and minimising the need for extensive individualised fine-tuning. TD3+BC modifies the established TD3 algorithm \cite{Fujimoto2018AddressingMethodsc}, introducing a behavioural cloning term in the temporal-difference update to discourage the selection of out-of-distribution actions \cite{Fujimoto2021ALearningc}. The policy, $\pi$ is formulated as:
\begin{equation}
    \pi = \argmax_{\pi} \hspace{-2mm} \mathop{\mathbb{E}}_{(s_i,a_i)\sim D} \hspace{-1mm} \left[ 
    \underbrace{\lambda Q(s_i, \pi(s_i))}_{\smash{\raisebox{-3mm}{\scriptsize \text{\shortstack{reinforcement \\ learning}}}}} 
    - \underbrace{(\pi(s_i) - a_i)^2}_{\smash{\raisebox{-3mm}{\scriptsize \text{\shortstack{behavioural \\ cloning}}}}} 
    \right],
\label{eq:td_update}
\end{equation}
where $Q^\pi(s_i, a_i) = \mathbb{E} \left[ r(s_i, a_i) + \gamma V^\pi(s_i') \right]$ denotes the state-action value function, $s'$ denotes the next state, and \(\lambda\) is a coefficient dictating the strength of the behavioural cloning effect. This approach has previously demonstrated state-of-the-art performance in T1D management tasks \cite{Emerson2023OfflineDiabetesc,Beolet2023End-to-endControl}.

PAINT reframes blood glucose management as multi-objective optimisation task, balancing user preferences (i.e. adapting to lifestyle and incorporating management advice) with the universal T1D goal of minimising patient risk. To achieve this, TD3+BC was initially pre-trained using a safety-focused reward function, before being tuned using patient-generated preference data. The tuning procedure is performed with the constraint that agent actions are similar to those of a safety-focused priori policy $\pi_{\text{priori}}$ and included by modifying the behavioural cloning term in \cref{eq:td_update} to $(\pi(s_i) - \pi_{\text{priori}}(s_i))^2$. This ensures the policy does not dangerously deteriorate under the presence of adversarial or misguided patient feedback and provides an intuitive method for patient fine-tuning of the strength of their preferences by modifying $\lambda$. The full training procedure is described in detail in \autoref{fig:full_algorithm_overview}.

\subsection{Simulated Type 1 Diabetes Patient}

Evaluation of the flexible blood glucose controller was performed using simulated patients and feedback.

\subsubsection{Glucose Dynamics Environment} 
Experiments were performed using the UVA/Padova T1D simulator \cite{Xie2018SimglucoseV0.2.1}, which provides a testing environment for RL blood glucose controllers and is approved as an animal testing substitute in blood glucose controller development by the FDA \cite{DallaMan2014TheFeaturesb}. This environment simulates the metabolic system of a cohort of patients with T1D; taking scalar insulin doses as input actions, $a$ and outputting a state describing blood glucose, $g_t$ carbohydrate consumption, $C_t$ and prior insulin doses, $I_t$ for a given timestep, $t$. Termination is possible and occurs when blood glucose exits the 10 to 1,000 mg/dL range, at which life-threatening harm would occur. 

The state was selected from \citeauthor{Emerson2023OfflineDiabetesc} and modified to \cite{Emerson2023OfflineDiabetesc}:
\begin{equation}
\mathbf{s} = \left[g_t, \mathbf{g}, \mathbf{I}, \text{IOB}, \text{COB}, W, \overline{a} \right],
\end{equation}
where $\mathbf{g} = [\overline{g}_{t-30}, \dots, \overline{g}_{t-240}]$ and $\mathbf{I} = [\overline{I}_{t-30}, \dots, \overline{I}_{t-240}]$ represent the mean blood glucose and insulin levels over the prior four hours (representing the maximum duration of insulin and carbohydrate activity) at 30-minute intervals, $W$ is the patient weight and $\overline{a}$ is the mean basal action. IOB (insulin-on-board) and COB (carbohydrates-on-board) approximate the activity of insulin and carbohydrates in the body and were approximated as an exponential decay, as utilised in the real-world Loop insulin controller \footnote{Loop Documentation: \href{https://github.com/LoopKit/Loop}{https://github.com/LoopKit/Loop}}. 
A detailed explanation of these functions is given in the Appendix.

Three virtual patients were chosen for training and evaluation, representing the median patient by weight for each cohort (adult, adolescent and child). Demonstrations were generated using a PID controller, which represents a widely-used and state-of-the-art controller in commercial insulin dosing devices \cite{Thomas2022AlgorithmsOverview}. Mealtime insulin dosing is typically performed by the patient. Bolus doses, $B_t$ were modelled using a simple bolus calculator, as utilised in prior work \cite{Emerson2023OfflineDiabetesc}. Both the PID benchmark and bolus calculator are described in Appendix.

\subsubsection{Simulated Feedback}

Patient preference labels were simulated to provide a reproducible and easily modifiable dataset for prototyping. Reward sketching was performed manually, using pre-specified preference functions. Patient preference labels were created by applying the preference functions to continuous segments of the data, processing state-action pairs and outputting a continuous value between -1 and +1 for each sample (-1 representing strong opposition to the patient's goal and +1 representing strong agreement).

The verifiably safe policy, as described in \autoref{sec:safety_orl}, was generated using TD3+BC trained using the Magni risk function, which is an established metric of glycaemic risk \cite{Emerson2023OfflineDiabetesc,Fox2020DeepControlb,Beolet2023End-to-endControl} and incentivises blood glucose in the target healthy range \cite{Kovatchev1997SymmetrizationApplicationsb}:
\begin{equation}
\label{eq:magni_risk}
r_{\text{magni}}(g_t) = 10 \cdot \left(c_1 \cdot \left(\log(g_t)^{c_2} - c_3\right)\right)^2,    
\end{equation}
where $c_1 = 3.5506$, $c_2 = 0.8353$, and $c_3 = 3.7932$ are constants. The Magni risk function is used throughout this work as T1D-specific metric of reward. 

\section{Experiments}

Evaluation of the flexible RL controller was divided into three core areas:
\begin{itemize}
    \item \textbf{Improving State-of-the-art} - replicating and enhancing the features of current non-RL based controllers.
    \item \textbf{Leveraging Patient Expertise} - exploring the method's utility for incorporating personalised patient knowledge for better control.
    \item \textbf{Real-World Feasibility} - assessing practical difficulties which could make real-world integration challenging.  
\end{itemize}

All RL algorithms in this work, unless specified otherwise, were trained on 100,000 samples (approximately six months) of pre-collected blood glucose data per patient, collected over continuous intervals of ten-days. Similarly, 10,000 samples (approximately three weeks) were labelled using the simulated patient preference functions. Each experiment was repeated over three random seeds, in which the reward model, verifiability safe policy and safety-constrained controller were re-trained. Evaluation was performed across ten-day continuous periods, and repeated five times. Reported results represents the median value across the full cohort of patients and their individual seeds.  

The hyperparameters of both the offline RL and reward learning algorithms were kept constant for all experiments and described in the Appendix. This was performed to ensure consistency with the real-world setting, where modified hyperparameters would require real-world testing on the patient and therefore would expose the user to unnecessary risk. The single parameter, $\lambda$ was modified between experiments, mimicking the real-world use of the algorithm, where the patient would tune this parameter to control the strength of their preferences.     


\subsection{Improving on State-of-the-Art}
\label{sec:sota_improv}

Current non-RL based blood glucose controllers are flexible to user preferences, but are not able to achieve comparable performance to RL-based controllers in simulated tasks \cite{Emerson2023OfflineDiabetesc,Beolet2023End-to-endControl}. The ideal controller would allow easy modification, without sacrificing performance or safety. 

\subsubsection{Setting Blood Glucose Targets}

\begin{table}[t!]
\centering
\begin{tabular}{cccc}
\toprule
\thead{Blood Glucose \\ Target (mg/dL)} & \thead{PID Reward} & \thead{RL Reward} & \thead{Target \\ Achieved} \\ 
\midrule
100 & -34,490 & -33,420 & \textcolor{green}{\ding{51}}\\ 
120 & -28,430 & -22,800 & \textcolor{green}{\ding{51}} \\ 
140 & -23,320 & -17,560 & \textcolor{green}{\ding{51}} \\ 
None & -20,730 & -20,680 & \textcolor{green}{\ding{51}} \\ 
160 & -18,100 & -18,750 & \textcolor{green}{\ding{51}} \\ 
180 & -25,220 & -22,854 & \textcolor{green}{\ding{51}} \\ 
200 & -35,940 & -34,190 & \textcolor{green}{\ding{51}} \\ 
\midrule
Mean & -26,700 & -24,320 & \textcolor{green}{\ding{51}} \\
\bottomrule
\end{tabular}
\caption{Comparison of PAINT to the PID benchmark for different blood glucose targets. Reward is measured via the Magni-risk function, as specified in \cref{eq:magni_risk}. The target is achieved if the mean blood glucose value of the RL agent is within $\pm$ 5 mg/dL of the median PID value. The RL agent meets the target and achieves lower risk in almost all instances.}
\label{tab:bg_targets}
\end{table}

PID controllers can be adapted to preference and expertise by modifying the target blood glucose level, representing the controller's equilibrium point. Setting a lower target can enable a more aggressive control strategy, but raises the risk of low blood glucose events. Similarly, raising the target often results in less risk to the patient, but a reduced performance. A successful RL controller should be able to replicate this basic functionality. \autoref{tab:bg_targets} shows that PAINT achieves greater reward across almost all target instructions and meets the target blood glucose value in each instance. PAINT demonstrates a competency in performing multi-objective optimisation, meeting patient instructions while developing lower risk control strategies.  

\subsubsection{Adapting to Common Management Goals}

Patients may modify their controller parameters to satisfy a broader T1D goal. Achieving these goals requires the patient to have an in-depth understanding of their condition and how their actions and environment influence it. They must also know how to adjust their control parameters to produce the desired response.

This experiment explores the potential of PAINT to achieve critical management goals without requiring patient insight on how to achieve them. This task uses three common example objectives:
\begin{itemize}
    \item \textbf{Increase time-in-range (TIR)} - improve the percentage of time spent in the target range, 70 to 180 mg/dL.
    \item \textbf{Reduce time-below-range (TBR)} - minimise the percentage of time below the target range, in the high-risk hypoglycaemic region ($<$ 70 mg/dL).
    \item \textbf{Reduce glycaemic variability (CoV)} - minimise the coefficient of variation in blood glucose.
\end{itemize}
Three preference functions were created representing each of these goals, respectively, and are presented in the first column of \autoref{fig:reward}. These functions embody simple labelling strategies, such as rewarding blood glucose in healthy glucose range (Raise TIR), less aggressively penalising high blood glucose (Lower TBR), and punishing blood glucose uniformly from the target (Lower CoV). As before, the performance of PAINT was compared to the PID benchmark. Optimisation of patient preference was performed for the PID benchmark by modifying the blood glucose target at 5 mg/dL graduations over the range of 100 mg/dL to 200 mg/dL and selecting the parameters best meeting the goal metric, while ensuring the median reward did not exceed -35,000.

\begin{table}[t!]
\centering
\begin{tabular}{p{7em}ccc}
\toprule
\thead{Patient Objective} & \thead{PID Reward} & \thead{RL Reward} & \thead{Goal \\ Achieved} \\ 
\midrule
Raise TIR & -34,489 & -24,180 & \textcolor{green}{\ding{51}}\\ 
Lower TBR   & -23,320 & -23,420 & \textcolor{green}{\ding{51}}\\ 
Lower CoV   & -25,220 & -21,480 & \textcolor{green}{\ding{51}}\\ 
\midrule
Mean & -27,680 & -23,030 & \textcolor{green}{\ding{51}} \\
\bottomrule
\end{tabular}
\caption{Comparison of PAINT to the PID benchmark in achieving the common objectives. Reward is measured via the Magni risk function, as specified in \cref{eq:magni_risk}. A goal is achieved if the metric change exceeds the PID benchmark. PAINT improves the metric beyond the the PID in all examples, while overall reducing patient risk.}
\label{tab:common_goals}
\end{table}

\autoref{tab:common_goals} shows PAINT achieves greater improvements across each metric compared to the PID benchmark, also obtaining reduced patient risk. Most significantly, achieving these goals requires no patient expertise in how to achieve the goal, only that they can identify goal states within their pre-collected data.

\subsection{Leveraging Patient Expertise}
\label{sec:patient_exp}

Patients' insights and experience in managing T1D could potentially enhance glucose control, improving event-specific management via human feedback. This experiment introduces several T1D case studies, illustrating how human expertise can be used improve control and meet patient goals. 

\subsubsection{Case Study 1: Regular Mealtimes}

Knowing a patient's meal schedule may allow PAINT to pre-empt meals, enabling earlier insulin administration and reducing glucose spikes from insulin delays. To include this, patient labelling was simulated using the preference function $r(a_t = \text{meal}(t) \cdot \text{a}_t^2$, where $\text{meal}(t) = 1$ in the two hours preceding the mean time a meal occurs and is zero otherwise. This function promotes higher insulin doses in the two hours leading up to a meal. 

\begin{figure}[t!]
    \centering
    \begin{subfigure}{0.53\linewidth}
        \centering
        \includegraphics[width=\textwidth]{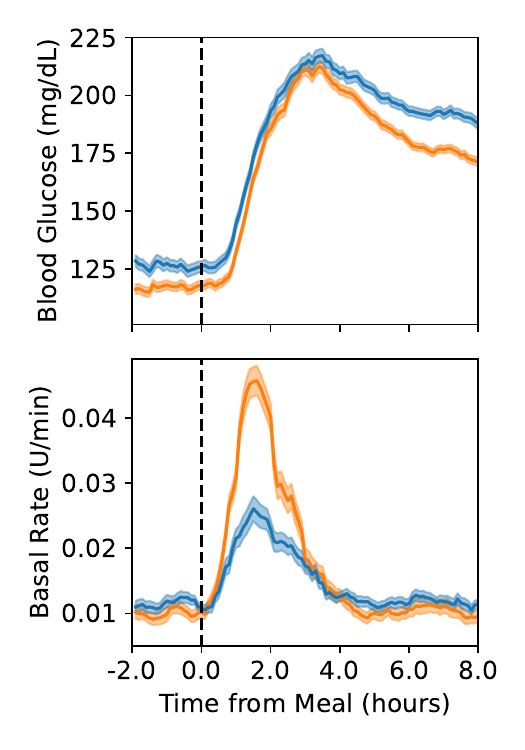}
        \caption{Regular mealtimes}
        \label{fig:meal_time}
    \end{subfigure}
    \hspace{-4mm}
    \begin{subfigure}{0.485\linewidth}
        \centering
        \includegraphics[width=\textwidth]{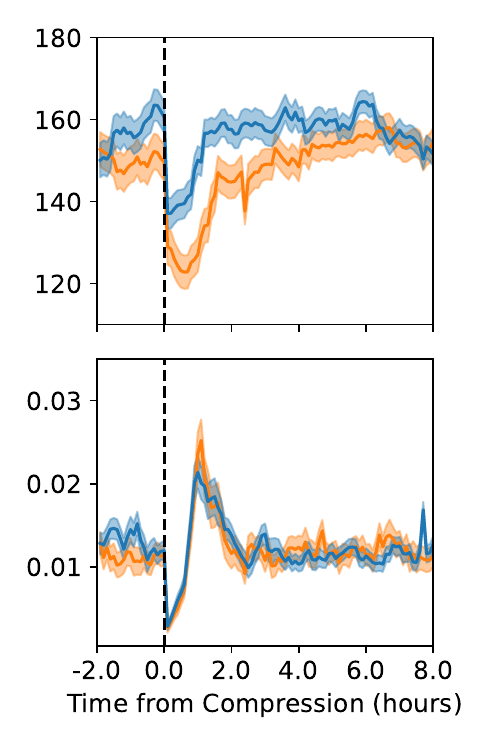}
        \caption{Compression lows}
        \label{fig:compression_low}
    \end{subfigure}
    \caption{Insulin dosing behaviour \textbf{\textcolor{orange}{with}} and \textbf{\textcolor{dodgerblue}{without}} human feedback: a) incentivising pre-emptive insulin dosing prior to regular mealtimes and b) penalising erroneous drops in insulin during compression lows. Human feedback results in a 10\% increase in TIR post-meal consumption and a 30\% increased insulin dose during compression lows (more closely matching optimal behaviour). Error bars represent the standard error.}
\label{fig:case_study_exp}
\end{figure}

\autoref{fig:meal_time} shows more aggressive dosing behaviours at meal time when utilising human feedback, improving the median TIR post-meal from 33.2\% to 44.2\%, and lowering the median patient risk ($-25,470$ to $-21,510$). The more aggressive strategy results in an increase in the max basal rate post-meal from 0.026 U/min to 0.046 U/min. Despite the improvements, the insulin dosing behaviour does not fully adhere to instructions, instead increasing the insulin dose post-meal rather than in the two hours prior. This may result from a fundamental limitation of offline RL, where the diversity of the demonstrator strategy limits the ability of the agent to meaningfully extrapolate to novel policies \cite{Nguyen-Tang2023OnBeyond}. The PID demonstrator shows significantly greater action standard deviation in the two hours following a meal than before (0.011 U/min to 0.022 U/min), which may justify the difficulty in modifying pre-meal strategy. This may be less significant in the real-world, as control strategies will need greater diversity to account for the more many factors influencing blood glucose dynamics.

\subsubsection{Case study 2: Compression Lows}

CGMs can give falsely low readings when they are compressed, such as during sleep, leading to \textit{compression lows} \cite{Idi2022Data-DrivenSensors}. These are characterised by sharp drops in blood glucose levels, followed by a rapid rebound. Blood glucose controllers frequently misinterpret these fluctuations as genuine events, leading to reduced insulin dosing. This, in turn, causes actual glucose levels to rise and contributes to increased glycaemic variability. If a controller could recognise these events and maintain normal dosing, it could improve control. To investigate this, the UVA/Padova simulator was adapted to simulate compression lows, as outlined in \citeauthor{Emerson2023OfflineDiabetesc} \cite{Emerson2023OfflineDiabetesc}. A preference function of, $r(a_t) = \text{comp}(t) \cdot \text{a}t^2$, was utilised, where $\text{comp}(t)$ equals 1 if $\left|g_t - g{t-5}\right| > 15$, encouraging higher insulin doses during rapid blood glucose drops.  

\autoref{fig:compression_low} demonstrates a significant increase in insulin dosing post-compression when utilising human feedback (0.024 U/min to 0.032 U/min). This is coupled with a reduction of CoV from 32.7\% to 31.1\% in the eight hours following. PAINT also experiences a modest reduction in risk from -23,300 to -22,210. Despite the observed change, the minimal change in reward may result from the RL controller already compensating for the compression low and is nearing the performance limit, as has been observed in prior work \cite{Emerson2023OfflineDiabetesc}. PAINT does not completely remove the erroneous drop in insulin, but clearly compensates for the drop more aggressively than before. As in case study 1, this may result from training dataset diversity data with standard deviation in the one hour after a compression low being almost half of that in the next (0.016 U/min compared to 0.028 U/min). 


\subsection{Real-World Feasibility}

PAINT's deployment success will also depend on its robustness to real-world challenges. 

\subsubsection{Sample Efficiency}

\begin{figure}[t!]
    \hspace{-5mm}
    \centering
    \includegraphics[width=1.05\linewidth]
    {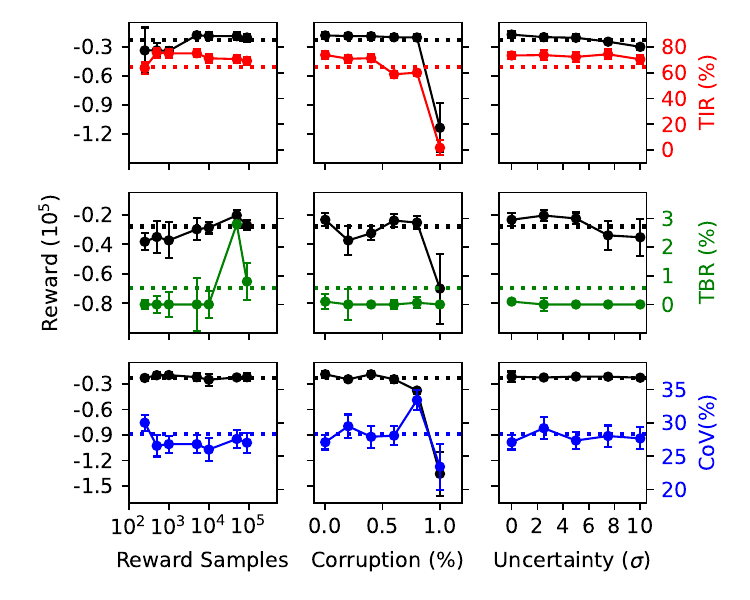}
    \caption{Robustness of PAINT to real-world challenges compared across three common T1D goals. PAINT is shown to be surprisingly effective under real-world constraints; achieving competitive results with $<$1,000 labelled samples (approximately 2 days of data), maintaining a performant agent with 80\% corrupted training data, and demonstrating marginal performance reductions with reward labelling noise, even up to 10$\times$ standard deviation, $\sigma$. The dotted lines act as a benchmark and indicate the parameter value without human feedback. Error bars describe the standard error.} 
    \label{fig:feasibility_study}
\end{figure}

Sample efficiency is important for deployment success, as patients are unlikely to benefit from PAINT if it is overly time-consuming. The number of reward labelled samples was varied from 250 (0.25\% of total samples) to 90,000 (90\%). The leftmost column of \autoref{fig:feasibility_study} shows overall that a greater number of labelled samples resulted in higher reward for the three objectives. Surprisingly, there is slight deterioration in TIR, TBR and CoV with greater percentages of reward labelled samples (most notably at 50,000 samples for TBR). This may imply that preference strength is partly dependent on the number of labelled samples, as at greater quantities the safety-constraint appears to be more strongly represented in the the tuned policy. Encouragingly, the agent improves TIR, TBR and CoV beyond the benchmark with as little as 1,000 reward labelled samples (approximately 2 days of data). This is encouraging for its real-world application, as greater sample-efficiency poses a reduced burden to the patient. 

\subsubsection{Incorrect Labelling}

Patients are likely to make errors when labelling reward, which may cause performance to deteriorate. Incorrect reward labelled samples were introduced into the training dataset. These samples used the negative of the original patient preference function, as to mimic extreme labelling errors. The central column of \autoref{fig:feasibility_study}, illustrates that PAINT show unexpectedly strong performance with incorrect reward labels. Only experiencing significant deterioration after 100\% of the data possesses the incorrect reward. This is positive for the safety of PAINT, as only under a very strong incorrect reward signal will the algorithm's performance diminish. As before, this effect could be attributed to offline RL's inherent pessimism, which has been observed in prior works to constrain the agent's policy \cite{Shin2023BenchmarksLearningb,Li2023SurvivalLearning}. Maximising TIR, minimising TBR and reducing CoV are likely to be more aligned with the objective of the training demonstrator and consequently, it is likely easier to extrapolate the agent's policy to achieve these goals. This highlights the importance of using verifiably safe training strategies for overall algorithmic safety. 

\begin{figure}[t!]
    \centering
    \includegraphics[width=0.9\linewidth]
    {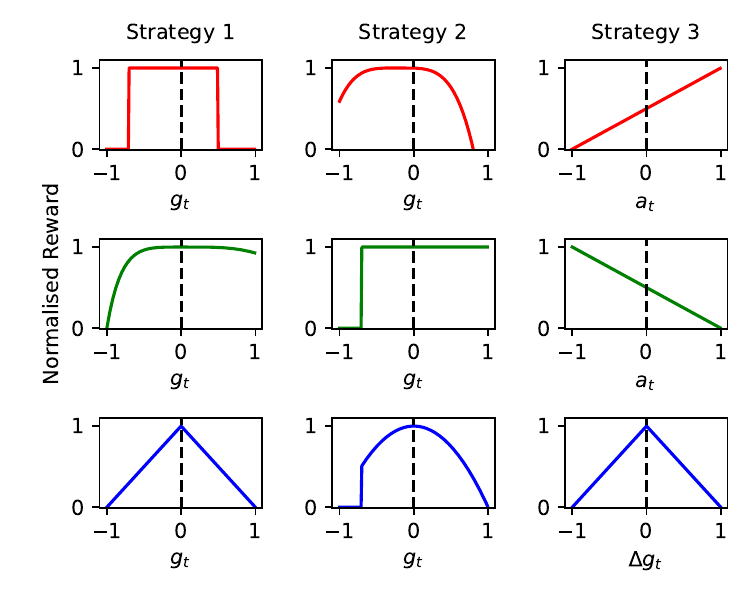}
    \vspace{-4mm}
    \caption{Nine different reward labelling strategies. Three for \textbf{\textcolor{red}{improving TIR}}, \textbf{\textcolor{mplgreen}{reducing TBR}}, and \textbf{\textcolor{blue}{minimising CoV}}. $g_t$, $a_t$, and $\Delta g_t = \left(g_t - g_{t-30}\right)$ are the blood glucose, basal action, and successive difference in blood glucose, respectively. The functions are described mathematically in the Appendix.}
    \label{fig:reward}
\end{figure}

\subsubsection{Labelling Uncertainty}

In addition to incorrect labelling, PAINT should also allow for imprecision in labelling. This was introduced in the form of Gaussian noise, added to the reward labels in multiples of the reward datasets standard deviation, $\sigma$. From the rightmost column of \autoref{fig:case_study_exp}, PAINT demonstrates high resilience to labelling uncertainty, with all results in error of the non-preference tuned benchmark. T1D metric scores do deteriorate with greater uncertainty, but still achieve competitive results for even 10$\times$ the reward standard deviation. This suggests there is a relatively large margin of error when performing reward labelling.  

\subsubsection{Diverse Labelling Strategies}

Patients are likely to adopt diverse labelling strategies, even for achieving the same objective. Three variations of reward labelling strategies were developed for each of the previously described T1D goals to test the versatility of PAINT. These strategies show variation in the variables they consider, functional complexity, and the extent to which they incorporate patient expertise and are visually represented in \autoref{fig:reward}.

\begin{table}[t!]
\centering
\begin{tabular}{p{4em}cc}
    \toprule
    Strategy & \thead{Reward} & \thead{Metric Change (\%)} \\
    \midrule
    TIR 1 & -17,560 & +3.8 \\
    TIR 2 & -20,430 & +4.0 \\
    TIR 3 & -24,180 & +6.4 \\
    \midrule
    TBR 1 & -23,420 & -0.6 \\
    TBR 2 & -22,310 & -0.6 \\
    TBR 3 & -28,480 & -0.6 \\
    \midrule
    CoV 1 & -21,480 & -1.3 \\
    CoV 2 & -20,160 & -1.9 \\
    CoV 3 & -28,640 & -0.5 \\
    \bottomrule
\end{tabular}
\caption{Performance of PAINT with the different reward labelling strategies presented in \autoref{fig:reward}, ordered sequentially from left to right. All reward labelling strategies elicit the metric change intended by the patient instruction, with small variation in patient risk.} 
\label{tab:diverse_reward}
\end{table}

\autoref{tab:diverse_reward} demonstrates that all labelling strategies successfully change the T1D metric as intended. The three methods achieving the lowest reward, label actions $a_t$ or consider more complex functions of the state, such as $\Delta g_t$. The deterioration with action labelling likely results from ambiguity in the intended goal state, as this labelling strategy highlights the path to the goal than the goal itself. Similarly, the complexity of $\Delta g$ may mean more samples are required to infer the goal state. Of note, PAINT is demonstrated to work well with binary labelling, as indicated by the TIR 1 and TBR 2 strategy in \autoref{tab:diverse_reward}. This method is particularly time-efficient and would enable rapid labelling of large numbers of examples.

\section{Conclusion}

This work presents a novel method for training flexible RL policies from human feedback. This method establishes an original approach for capturing diverse patient preferences and fine-tuning an offline RL controller to satisfy the constraints of a multi-objective task. This approach demonstrates flexibility and performance in excess of current control benchmarks, providing a simple method for users to achieve complex goals. This methods shows robustness to real-world challenges, such as sample efficiency, diverse labelling, and labelling errors, highlighting its potential for real-world evaluation in future work. 


\bibliography{ijcai25}
\bibliographystyle{named}


\section{Appendix}

\subsection{Data and Code Availability}

Code will be made available on acceptance and contains the configuration files necessary to replicate the training dataset and run the experiments. Training was parallelised across four NVIDIA GeForce RTX 2080 Ti GPUs, with the full training process taking a approximately 30 minutes per run.

\subsection{Hyperparameters of PAINT}
\label{sec:hyperparams}

The hyperparameters for PAINT are presented in \autoref{tab:hyperparams}. TD3+BC was modified to incorporate n-step Q-learning, as this was necessary to model the extended effects of insulin and carbohydrates. The presented hyperparameters were fixed across all the presented experiments. 

\begin{table}[t]
\centering
\begin{tabular}{@{}p{2cm}ll@{}}
\toprule
& Hyperparameter & Value \\
\midrule
\multirow{4}{*}{\centering Training} & Optimiser             & Adam            \\
                                             & Mini-Batch Size       & 256             \\
                                             & Training Epochs       & 300             \\
                                             & Tuning Epochs         & 150             \\
\midrule
\multirow{4}{*}{\centering Actor }    & Actor Features        & 256             \\
                                             & Actor Layers          & 2               \\
                                             & Actor Learning Rate   & 3e-4            \\
                                             & Actor Activation      & ReLU            \\
\midrule
\multirow{4}{*}{\centering Critic}   & Critic Features       & 256             \\
                                             & Critic Layers         & 2               \\
                                             & Critic Learning Rate   & 3e-4            \\
                                             & Critic Activation     & ReLU            \\
\midrule
\multirow{8}{*}{\centering General} & $N$-steps             & 10              \\
                                             & Reward Scale    & 1,000            \\
                                             & Training Alpha        & 2.5             \\
                                             & Policy Update Frequency & 2             \\
                                             & Policy Noise          & 0.2             \\
                                             & Policy Noise Clipping & (-0.5, 0.5)     \\
                                             & Discount Factor       & 0.999           \\
                                             & Target Update Rate    & 5e-3            \\
\midrule 
\multirow{6}{*}{\centering Reward} & Learning Rate & 4e-5    \\ 
        & Mini-Batch Size & 128   \\ 
        & Early Stopping & True   \\ 
        & Training Epochs & 500  \\ 
        & Network Features & 256  \\
        & Network Layers & 3     \\
        & Reward Bins  & 10      \\
\bottomrule
\end{tabular}
\caption{Hyperparameters for the Safety-Constrained Offline RL component of PAINT.}
\label{tab:hyperparams}
\end{table}

\subsection{Formulation of IOB and COB}
\label{sec:iob_formulation}

Insulin-on-board (IOB) and carbohydrates-on-board were approximated using the following equations derived from the open-source Loop insulin dosing controller:
\begin{equation}
\text{activity}(t) = \frac{S}{\tau^2} \cdot t \cdot \left( 1 - \frac{t}{t_d} \right) \exp\left(-\frac{t}{\tau}\right),
\end{equation}
where $S$ is
\begin{equation}
S = \frac{1}{1 - a + (1 + a) \exp\left(-\frac{t_d}{\tau}\right)},
\end{equation}
with $a = \frac{2 \tau}{t_d}$ and $\tau$ defined as
\begin{equation}
\tau = \frac{t_p \left( 1 - \frac{t_p}{t_d} \right)}{1 - \frac{2 t_p}{t_d}}.
\end{equation}
The parameters $t_p$ (peak time) and $t_d$ (duration) were set to $t_p = 55$ min, $t_d = 240$ min for IOB, and $t_p = 40$ min, $t_d = 210$ min for COB, to match the activity profiles from the UVA/Padova simulator. IOB and COB were then computed by taking the summation of all prior insulin and carbohydrates activities over $t_d$. 

\subsection{PID Benchmark}
\label{sec:PID_benchmark}

PID algorithm operates according to:
\begin{equation}
\label{eq:pid_controller}
    a_t = k_p \cdot \left( g_\text{targ} - g_t \right) + k_i \cdot \sum^t_{t'=0} \left(g_{t'} - g_\text{targ} \right) + k_d \cdot \left( g_t - g_{t-1} \right),
\end{equation}
where $g_{targ}$ is the target blood glucose value and $k_p$, $k_i$ and $k_d$ are parameters tuned to the patient. Gaussian noise was introduced to the PID parameters to represent patient ambiguity in determining the optimal PID parameters. A small amount of Ornstein-Uhlenbeck noise was also added to controller actions to mimic the inherent uncertainty of real-world systems. 

\subsection{Bolus Calculator}
\label{sec:bolus_calculator}

The equation to compute meal-time insulin, $B_t$:
\begin{equation}
    B_t = \frac{c_t}{\text{CR}} + \left( \sum^{N}_{t'=0} c_{t-t'} = 0 \right) \cdot \frac{g - g_{\text{targ}}}{\text{CF}},
\end{equation}
where $g_t$ is blood glucose level, $g_{\text{targ}}$ is the target blood glucose, $c_t$ is the quantity of consumed carbohydrates, $CR$ and $CF$ are patient-specific parameters. A uniform uncertainty in carbohydrate quantity, $c_t$ was included to the difficulty of  approximating its content in real-world meals. 


\subsection{Patient Preference Functions}
\label{sec:preference_funcs}

\begin{table*}[t]
\centering
\renewcommand{\arraystretch}{1.5}
\begin{tabular}{p{1cm}p{6cm}p{7cm}}
\toprule
Strategy & Equation & Rationale \\
\midrule
TIR 1 & $ r(g_t) = -\left(g_t - 125\right)^4$ & Incentivises blood glucose being in a tight region surrounding 125 mg/dL. \\ 
TIR 2 & $r(g_t) =\begin{cases} 1 & \text{if } 70 < g_t < 180, \\
        0 & \text{otherwise.} \end{cases}$ & Reward is only given when blood glucose is in the target range of 70 to 180 mg/dL. \\ 
TIR 3 & $ r(a_t) = a_t$ & Encourages more insulin to be given, resulting in a more aggressive strategy. \\ 
\midrule 
TBR 1 & $ r(g_t) = \left(r_{\text{magni}}(g_t)\right)^2$ & Modified magni risk function to punish low blood glucose values more than before. \\ 
TBR 2 & $ r(g_t) = \begin{cases} 1 & \text{if } g_t > 70, \\ 0 & \text{otherwise.} \end{cases}$ & Reward is given only if patient is not in the low blood glucose region (70 mg/dL). \\ 
TBR 3 & $ r(a_t) = -a_t$ & Encourages less insulin to be given, resulting in higher blood glucose levels on average. \\ 
\midrule
CoV 1 & $r(g_t) = -|g_t - 144| $ & Constrain blood glucose more closely to a target of 144 mg/dL. \\ 
CoV 2 & $ r(g_t) = \begin{cases} r(g_t) = \left(g_t - \bar{g}\right)^2 & \text{if } 70 < g_t \\ 0 & \text{otherwise.}\end{cases}$ & Peanalise deviations from the mean and severely penalise hypos. \\ 
CoV 3 & $ r(\Delta g) = -|\Delta g| $ & Penalise abrupt changes in blood glucose over 30-minute periods. \\ 
\bottomrule
\end{tabular}
\caption{The individual reward strategies, accompanied by the rationale behind each selected equation. $g_t$ is the blood glucose measurement in timestep $t$, $a_t$ is the basal insulin action, $\Delta g = \left(g_t - g_{t-30}\right)$ is the difference in blood glucose over a 30-minute interval, $\bar{g}$ is the mean blood glucose, and $r_{\text{magni}}$ is the Magni risk function.}
\label{tab:reward_strategies}
\end{table*}

The simulated patient preference functions used in this work are presented in \autoref{tab:reward_strategies}. The rationale behind each selected preference function is given, as to represent the intention of a person with type 1 diabetes.

\end{document}